\documentclass[twocolumn,10pt]{asme2ej}

\usepackage{graphicx}
\usepackage{amsmath}
\usepackage{hyperref}   % to set up hyperlinks
\hypersetup{
	colorlinks=true,
	linkcolor=blue,
	citecolor=blue,
	urlcolor=blue,
}
\usepackage[square,numbers]{natbib}

\title{Deployment of long distance multi-moving
robots for underground pipe inspection}

\author{\qquad Weiyao Lai} \author{\qquad Wei Xu}
\author{Marc Bernhard}

\begin{document}

\maketitle    

%%%%%%%%%%%%%%%%%%%%%%%%%%%%%%%%%%%%%%%%%%%%%%%%%%%%%%%%%%%%%%%%%%%%%%
Blueprint of an in-pipe climbing robot that works with sharp transmissions to study complex line relationships. Standard wheeled/happening pipe climbing robots tend to slide when exploring  pipe turns. Instruments help achieve a very distinct delay sequence in which the robot slides and drags as it progresses. The proposed transmission joins the  farthest ground plane of the standard two-output transmission. This opens up a substantial time for 3 output transmissions. This instrument takes into account the force exerted on each track within the line relation to specifically alter the robot's track speed, unlocking the key to fine control. Deflection of the robot across pipe networks with different bearings and non-slip pipe bends demonstrate the integrity of the  proposed structure.

%%%%%%%%%%%%%%%%%%%%%%%%%%%%%%%%%%%%%%%%%%%%%%%%%%%%%%%%%%%%%%%%%%%%%%
\section{Introduction}

After all considerations, pipe networks are clearly used to move liquids and gases in stables and metropolitan areas. Most commonly, the lines are covered to comply with safety rules and avoid possible consequences. This makes line inspection and maintenance a real test. Covered lines are particularly good for fainting, using, starting scaling progress and interruptions, performing upgrades or damage that can affect amazing episodes. Various inspection robots have been proposed in the past to perform standard preventive assessments to avoid disasters. We have also demonstrated that bioactivated robots with crawlers, inchworms, running parts \cite{adriansyah2017Optimization} and screw drive \cite{ravina2010low} structures are suitable for different needs. Regardless, most of them use dynamic control technology to guide and move in the line. The reliance on the robot's course in the line added to the difficulty, and the robot was similarly vulnerable to slippage if joint control procedures were not included. Theseus \cite{ravina2016kit,choi2007pipe,fujun2013modeling} The robot series uses separate parts for the driving and driven modules, which are interconnected with different connection types. Each area changes or shifts the orientation for locking turns. In addition, amazingly engaging control exposes such robots  to sensory information, which is greatly appreciated.

With 3 modified modules, the pipe climbing robot is more stable and more comfortable. A previously proposed restricted line climber \cite{vadapalli2019modular,suryavanshi2020omnidirectional,vadapalli2021modular} used his three lanes made by robots similar to the MRINSPECT series. For such a robot, the speed of cornering was indicated in advance in order to properly control the three tracks. This was intended to have the robot compose convincing line transformations at specific locations independently of previously presented rates. In a guaranteed application, the robot course shifts past it, sliding into the track as it progresses. This obstacle can be addressed by controlling the robot using potentially working transmission parts. MRINSPECT-VI \cite{chang2017development,litopic} uses multi-colossal transfer parts to control the speed of  three modules. However, a central transmission system is used, in which he distributes the  work and speed to  three modules. This perspective caused the focal yield (Z) to rotate faster than her other two results (X and Y), making the Z yield actually affected by hatch unfolding. This is caused by the fact that the inevitable aftermath of the transfer does not fundamentally add fuzzy energy to the data. Other controls that have actually been proposed have yielded real results with respect to transmission that did not follow the common system.

Forwarding sees tedious consequences for entering dynamic relations and kills supported keys \cite{vadapalli2021design}. receive. This causes the robot to  slip and drag during the redevelopment of the system for robots. The transmission portion of the Line Climber also restores comfort by circulating line affiliation with less reliance on trusty controls. This improvement really shifts power and speed from a given mission to her three lanes of robots through the intricate movement of materials, of course considering the stack each lane experiences.

%%%%%%%%%%%%%%%%%%%%%%%%%%%%%%%%%%%%%%%%%%%%%%%%%%%%%%%%%%%%%%%%%%%%%%
\section{Robot's Development}

The robot's odd cornerless central package houses three modules separated from each other. Gears machined inside the robot drive three chains through drive sprockets. Prepare yield-related obliques. A calibrated perspective of the  instrument  is revealed in the previous subchapter. Each module has rails and  openings to move the four connections. Modules are connected using direct springs attached to links (or shafts). H. Springs between module and robot body are pushed radially outward. Customizations are reference constructs that limit module improvement beyond the  OK endpoint. When the robot is sent to the line, it passes through a detachment with spring-loaded rails and is pushed against a partition inside the line. This gives the central consent for the robot to move. Similarly, each module in the robot may be unevenly packed per point. The rails attached to the modules are pressed against the bulkhead inside the line, giving the robot great freedom of movement. There are drive wheels that convert rotational motion from a  specific starting position into translational reinforcement of  the robot. The Dividers Squash game plan typically includes his 4 direct links to 3 modules. The module has an opening through which the straight link (or shaft) passes. They show the progress of the arc head's modules  from the rotation of the robot.

\begin{figure}[ht!]
\centering
\includegraphics[width=3.1in]{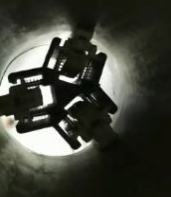}
\caption{\footnotesize Multi-moving robot}
\label{1}
\end{figure}

The gearbox is a huge part of the proposed robot. The instrument contains an information item, 3 transfers, 3 open transfers with 2 inputs and 3 results. The evaluation of the gear is represented by the rotation of the center of gravity of the robot body. The three appear consecutively around the information, placed between any two. They are mounted radially in the center. The only explicit aftermath of progress involves gear parts such as ring gears, machine gear and helical pinion deals, ring gear connections, and helical pinion and pinion gear battles. The machine's side gears mesh with  nearby side pinion gears to transfer power and speed  to the cover. Worm gear information also progresses at the same time. Each 2-input gearbox then evolved into a 2-input gearbox cover depending on the stuck conditions experienced by the various side sprockets. With the additional improvement provided by the side sprockets, they come in two. The six shows work together to facilitate the progression from Commitment to Three Outcomes. When different weights are generated, the side sprockets exchange different weights to the side sprockets. Transmission rate information is targeted in this state.

Arguably, all side sprockets experience weight that is indistinguishable when underestimated or stackless and spins at typical speeds and forces. Fanning out, each result shares a bewildering energy with the data. Moreover, the results share equally weak energies with each other. This means that when divided into loads, one of the results affects the other two, but they are not perturbed. Results generally do not work for many of the steps shown if there is no load or close weight to return to the results. If the resulting load is changing, the part will process the result at line speed. In any case, if one of the outcomes operates at the exchange rate and the other two encounter proper weights, the two outcomes operate with muddy weights in a chaotic procedure. The robot expects the part to move in a straight line pushing three lanes at breakneck speed. In either case,  the gearing changes the orbital speed of the robot while simultaneously moving within the line winch. The objective is the compelling goal that the trajectory corresponding to all long-range bits rotates faster than the trajectory corresponding to all long-range bits. Meet the limited distance. Please suggest \cite{vadapalli2021modular,9635853} for additional diagrams, plans and information.

\begin{figure}[ht!]
\centering
\includegraphics[width=3.1in]{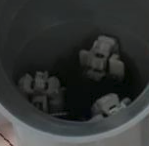}
\caption{\footnotesize Sectional View}
\label{1}
\end{figure}

%%%%%%%%%%%%%%%%%%%%%%%%%%%%%%%%%%%%%%%%%%%%%%%%%%%%%%%%%%%%%%%%%%%%%%
\section{Design Specifications}

Kinematic diagrams show the affiliations and joints of the instruments. The  kinematic and dynamic conditions of the transmission are handled using a bond outline model. Explicit speed information  from the engine is mandated according to his three ring gears in a two-way transmission. They rotate at  positive velocities with poor properties. Generally this is very important. For example, overcast is good for gear engagement. In addition, a two-way gearbox works, where the specific speed of the ring gear corresponds to the exorbitant speed of the  two side pinions for each season. These two side cross gears can rotate at different speeds while perceiving small forces. \cite{deur2010modeling}.

\begin{equation}
\arctan (\frac{AO}{OC}) = \arcsin (A^{'}C^{'}),
\label{1}
\end{equation}

From the security diagram model, we can decipher three gears spinning at infinite speed. Taking all factors into account, infinite power is a multiple of the problem level, for example, information  speed,  information power, and commitment to ring gear. Moreover, in a two-input gearbox, a certain speed of the  ring gear is standard and essential for the bold steps of the  two side pinions. These two side machine gears can rotate at different speeds while maintaining indistinguishable forces. Then place the test rate as a side machine gear. Auxiliary mechanical gears are associated with the undisputed goal of not having broad improvements between the pinion gears of the proportional pair. Insert (\ref{1}). As a rule, you get an  exact rate. The association from results to side materials meanders to meet the brisk speed requirement for committing to results. Due to the  difficult speed from the plate pinion  to the side machine gear, it achieves the ratio of data and side stop gear. Similarly, the results are not entirely separate from the lonely cog steps to create the Side Stuff Alliance results. Checking both relationships satisfies a special speed constraint to commit to the result. If a certain speed corresponds to information, then the filling level of the ring gear cog is the result, and the wise step is the result. Individually determined gears for lateral tamping gears. Therefore, the exact rate of results is affected by information flow and yield of by-products. Meanwhile, the force of the ring gear is as large as the force of the seeing side pinion. Separating the ring gear and side gear relationships as well as careful speed relationships for compliance commitments yields a ring gear partnership and a mapping between compliance forces.

\begin{equation}
(N + \mu N) \sin{(A^{'}C^{'})} = H
\label{2}
\end{equation}

Here the specific speed of  information is the filling level to the results of the gears of the ring machine and the steps of the results are executed. Yield is the individual exact speed of the side-stuffed gear. Therefore, the exact rate of results depends on the speed of the information light  and the yield of the pages. Meanwhile, the force of the ring gear is as large as the force of the seeing side pinion. Similar to the exact velocity relationship for compliance commitments, the mapping between compliance forces and information forces is obtained by separating the ring gear and side gear relationships using the ring gear mapping results.

\begin{equation}
K_s = \frac{ \mu N}{9}
\label{3}
\end{equation}

The condition \eqref{3} represents a transfer interest that gives similar motion credits in each of the three outcomes if undefined weights are encountered or  unconstrained. In any case, when the cloth encounters an obstacle through a rally point, its luminosity and  power change according to its apparent resistance. The resulting speed of the transmission is similar to the sprocket speed of the drive module. So the resulting transmission speed is not just converted to line speed. Thus, the robot's information velocity is essentially derived from results under ambiguous stacking conditions. Sprocket widths are generally reliable for 3 chains. In her study by  Ho Moon Kim et al., \cite{litopic} proposed a strategy for determining the specific velocities of  three tracks in a pipe curve. The direction in which the robot enters the pipeline in the strategy shown is a property of the line rotation, R is the width of the line wind melody improvement, and r is the line height. Also, the velocity of each track B and C is obtained. Robots are built into various locations in her OD module. Modified speed of the wind whistle, which is not always placed in the heading.

\subsection{Control}

Straight springs in the module allow the robot  to flexibly place curves with little difficulty. The highest grade for each module is $16 mm$. The opening of the module offers additional versatility, proving that upside-down printing is possible. This allows the robot to overcome obstacles and irregularities in wiring networks and inspect certified applications. The front  of the module is fully compressed regardless of whether the back is in the most critical extended state in the line. The most absurd and inconsistent loads allowed on a single module of a robot. So $\phi$ is the optimal point where modules can be packed unevenly. Derive \cite{vadapalli2021modular,9635853} for additional diagrams, overviews and information.

\begin{figure}[ht!]
\centering
\includegraphics[width=3.1in]{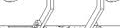}
\caption{\footnotesize Bend 1}
\label{asym(1)}
\end{figure}

\section{Experimentation}

Worked with Detour to research and maintain improvements to the robot's longest range in various strains. The same gives us more experience with the parts and orientation of the manufactured robot under reliable test conditions. From this point on, a dynamic multi-body spread was performed by switching to an improved reproduction model across range of motion limitations. Tracks were converted to roller wheels to reduce the number of  moving parts in the model and  reduce  computational weight. Each module houses 3 roller wheels of the better model. The contact fixation provided by the track to the line divider is thus reduced from 10 contact areas to 3 contact areas per module. Track speed and  module load for each track A, B and C have been reduced very favorably. Expansions have been made by introducing robots to her three undeniable headlines in modules, both  straight  and curved. The robot is highlighted within a line network supported by ASME B16.9  NPS 11 and Plan 40 standards. Diversions were performed for his four evaluation conditions of the line network, including vertical locales, elbow regions, horizontal pieces, and U-pieces for different orientations of the robot. Absolute distance for line structures. The distance traveled by the robot is not always set from the mixing point on the robot body, the specific distance of the track actually depends on the central roller wheels mounted on each module. So by killing the length of the robot  we get the actual robot course. Robot paths in vertical climbing and final level areas are concentrated by subtracting from their unitary length. The robot data  is given a solid astonishing speed, and improvements to the robot, including track speed, are incorporated into the reproduction.

\begin{figure}[ht!]
\centering
\includegraphics[width=3.1in]{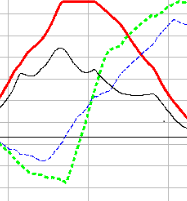}
\caption{\footnotesize Bend 2}
\label{asym(1)}
\end{figure}

On upward pieces and level quarters, the robot follows a straight path. Track therefore receives questionable weight in each of the three modules in both evaluations. Therefore, in case of doubt, the gearbox gives an indefinite value for the three tracks, such as the normal speed of the robot. The saw chain accelerates in flexure for propulsion. Similarly, each attribute is separated from the theoretical result by a level out rate (APE) wrench. This error outlines an exact measure of deviation from the speculative value \cite{armstrong1992error}. To get the base length  in vertical climbing, the robot guesses 0 to be 9 seconds. A robot's ability to climb a line despite gravity. First, the robot travels a distance of 350 mm in the hub level area. At a really unnoticeable level of distance. In the elbow region, the robot moves equidistant to the confluence of the  bits of the lines. The system will change the synthetic speed of the track as indicated by the burst, from a point that matches line conditions. In all three of his robot's directions, the outer module track moves faster and covers greater distances, while the inner module track reaches all bits at the closest distance than linewind bits. Because of this, it rotates with a significant delay. For tube reshaping, the detour speed of each track is individually displayed on the heartless at the midpoint of  the unchanged saw track speed. Approximate steps in each trace are found separately from their respective speculative rates using a common rate error (APE) search.

\begin{figure}[ht!]
\centering
\includegraphics[width=3.1in]{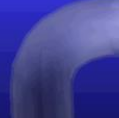}
\caption{\footnotesize Bend 3}
\label{asym(1)}
\end{figure}

On course, the outer modules (B and C) move at  normal speed, while the inner module (A) moves at a default speed of 33.62 mm/s. These qualities are combined with theoretical properties. Similarly, the line speed is also consistent with the reference value for improvement results. Similarly, line speed is an inspiration and different than the APE 2.5 build results. Whatever the strategy, straights or corners, ruined respect is utterly irrelevant and  can be credited with clear terms such as deterioration due to external factors. Additionally, unrelated velocity changes occur in the unfolded view. From 9s he is 24s, the robot accumulates distance in the elbow area, but it takes him 59s to enter the U-segment. The robot's ability to inspect the elbow. for shows that the outer modules (B and C) move with speed, whereas the inner module (A) he moves with one speed. These credits exist in a speculative nature. Likewise, the reproduction results have track speed and inspiration. The range of speeds derived from the detours is taken from the base of each chart to the most absurd speed of the highest speed. Please suggest \cite{vadapalli2021modular,9635853,saha2022pipe} for additional diagrams, diagrams and information.

The line speed improvement results in the  various rubrics are reconciled with the hypothetical results achieved in the region. From the simulations, we find that  the robot reliably explores  the entire route network using her 60s-introduced orientation. Result interface with  speculative quotes. The chain of the line climber has slippage and drag on the gears, so it can be improved without difficulty. Augmentation allows the robot to be seen anywhere without sliding or dragging, which reduces strain  on the robot and increases freedom of progression. The module's rails work in conjunction with the line's internal mass to provide balance while underway. A spring is embedded in the upstream position so that initially he is similarly pre-stacked by tension in each of the three modules. In a straight line, the robot travels the main length of the prestacked spring. As the robot gets closer to the elbow area and U-segment, the molded length of the inner and outer modules increases by 1.5mm. This variant demonstrates the wide adaptability of considering the module to pass through the lateral portion of the line width while progressing along the curve.

\section{Conclusion}
The robot gets nice transmission to control the robot clearly without any noticeable control. Transmission has confusing consequences for connecting energies. Its show is completely similar to the comfort of the standard two-result transmission. The age results support strong crosspoint bottoms of stunning line networks with  up to $^\circ$180 spots in various directions without slippage. Taking on the robot's transmission portion leads to the poignantly acknowledged end result of overcoming the slip and  the entire robot during a new turn of events. We are currently drawing  a model for considering the proposed plan.
%%%%%%%%%%%%%%%%%%%%%%%%%%%%%%%%%%%%%%%%%%%%%%%%%%%%%%%%%%%%%%%%%%%%%%

\bibliographystyle{asmems4}

\bibliography{asme2e}

\end{document}